\documentclass{article}
\usepackage{interspeech_04,amssymb,amsmath}
\def\reg{{\rm\ooalign{\hfil
     \raise.07ex\hbox{\scriptsize R}\hfil\crcr\mathhexbox20D}}}

\usepackage{graphicx}

\title{Unsupervised Topic Adaptation for Lecture Speech Retrieval}

\name{{\em Atsushi Fujii$^{\dagger}$, Katunobu Itou$^{\ddagger}$, Tomoyosi Akiba$^{*}$, Tetsuya Ishikawa$^{\dagger}$}}

\address{$^{\dagger}$ Graduate School of Library, Information and
  Media Studies, University of Tsukuba \\
  $^{\ddagger}$ Graduate School of Information Science, Nagoya University\\
  $^{*}$ Department of Information and Computer Sciences, Toyohashi
  University of Technology \\
  {\small \tt fujii@slis.tsukuba.ac.jp}
}

\newcommand{\eq}[1]{(\ref{#1})}

\begin{document}
\maketitle
\begin{abstract}
  We are developing a cross-media information retrieval system, in
  which users can view specific segments of lecture videos by
  submitting text queries. To produce a text index, the audio track is
  extracted from a lecture video and a transcription is generated by
  automatic speech recognition. In this paper, to improve the quality
  of our retrieval system, we extensively investigate the effects of
  adapting acoustic and language models on speech recognition. We
  perform an MLLR-based method to adapt an acoustic model. To obtain a
  corpus for language model adaptation, we use the textbook for a
  target lecture to search a Web collection for the pages associated
  with the lecture topic. We show the effectiveness of our method by
  means of experiments.
\end{abstract}

\section{Introduction}
\label{sec:introduction}

Given the growing number of multimedia contents available via the
World Wide Web and DVDs, retrieving specific information relevant to
user needs has become crucial. Because text is one of the most common
and effective methods to represent user needs, we proposed a
cross-media information retrieval (CMIR)
system~\cite{fujii:eurospeech-2003}, in which a user can submit text
queries to search an entire lecture video program for the relevant
segments.

Because oral presentations are usually organized based on text
materials, such as textbooks, a user first selects text segments
(e.g., keywords, phrases, sentences, and paragraphs) in a textbook
related to a target lecture. Then, a text query is generated
automatically from one or more selected segments. That is, queries can
be formulated even if the user cannot provide effective keywords. The
user can also submit additional keywords as queries, if
necessary. Finally, video segments are retrieved and presented to the
user.

To retrieve video passages in response to text queries, we extract the
audio track from a lecture video, generate a transcription by means of
automatic speech recognition (ASR), and produce a text index, prior to
system use.

In our previous work~\cite{fujii:eurospeech-2003}, we proposed a
method to adapt a language model for ASR to the topic of a target
lecture. We showed that our method improved the accuracy of ASR and
also improved the accuracy of CMIR significantly, by means of
experiments.

However, in the previous experiment the vocabulary size was limited to
20K, although a larger vocabulary size, such as 100K, has been used in
recent research. In addition, the contribution of acoustic model
adaptation on ASR has not been investigated. It may be argued that the
effects of adapting models are overshadowed by increasing the
vocabulary size. To answer this question, in this paper we extensively
investigate the effects of adapting the language and acoustic models
on ASR.

\section{Cross-media Retrieval System}
\label{sec:system}

\subsection{Overview}
\label{subsec:system_overview}

\begin{figure}[htbp]
  \begin{center}
    \leavevmode
    \includegraphics[height=2.9in]{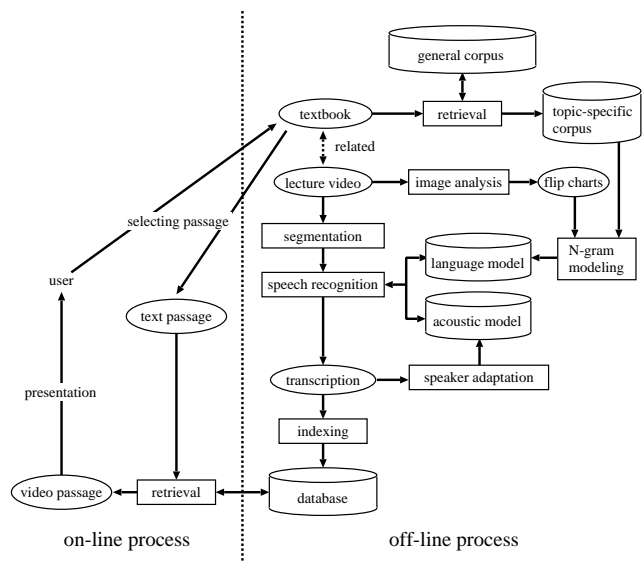}
  \end{center}
  \caption{Overview of our CMIR system.}
  \label{fig:system}
\end{figure}

Figure~\ref{fig:system} depicts the overall design of our CMIR system,
in which the left and right regions correspond to the on-line and
off-line processes, respectively. While our system is currently
implemented for Japanese, our methodology is fundamentally language
independent.  For the purpose of research and development, we
tentatively target lecture programs on TV for which textbooks are
published. We explain the basis of our system using
Figure~\ref{fig:system}.

In the off-line process, given the video data of a target lecture,
audio data are extracted and divided into a number of segments. Then,
a speech recognition system transcribes each segment. Finally, the
transcribed segments are indexed as in conventional text retrieval
systems, so that each segment can be retrieved efficiently in response
to text queries.

For speech recognition, we use two adaptation methods. To adapt speech
recognition to a specific lecturer, we perform unsupervised speaker
adaptation using an initial speech recognition result (i.e., a
transcription).  To adapt speech recognition to a specific topic, we
perform language model adaptation, for which we search a large corpus
for the documents associated with the textbook for a target
lecture. Then, the retrieved documents (i.e., a topic-specific corpus)
are used to produce a word-based N-gram language model.

We also perform image analysis to extract text (e.g., keywords and
phrases) from flip charts. These contents are also used to improve our
language model.  However, this method is beyond the scope of this
paper.

In the on-line process, a user can view specific video segments by
submitting any text queries, i.e., keywords, phrases, sentences, and
paragraphs, extracted from the textbook. Any queries not in the
textbook can also be used.  The current implementation is based on a
client-server system on the Web. Both the off-line and on-line
processes are performed on servers, but users can access our system
using Web browsers on their own PCs.

It should be noted that unlike conventional keyword-based retrieval
systems, in which users usually submit a small number of keywords, in
our system users can easily submit longer queries relying on
textbooks.  If submitted keywords are misrecognized in transcriptions,
the retrieval accuracy decreases.  However, long queries are robust
for speech recognition errors, because the effect of misrecognized
words is overshadowed by the large number of words correctly
recognized.

Because our focus in this paper is to investigate the accuracy of
speech recognition, in the following sections we elaborate only on
speech recognition and document retrieval used to obtain
topic-specific corpora.

\subsection{Speech Recognition}
\label{subsec:asr}

The speech recognition module generates word sequence $W$, given phone
sequence $X$. In a stochastic framework, the task is to select the $W$
maximizing $P(W|X)$, which is transformed as in Equation~\eq{eq:bayes}
through the Bayesian theorem.
\begin{equation}
  \label{eq:bayes}
  \arg\max_{W}P(W|X) = \arg\max_{W}P(X|W)\cdot P(W)
\end{equation}
$P(X|W)$ models the probability that the word sequence $W$ is
transformed into the phone sequence $X$, and $P(W)$ models the
probability that $W$ is linguistically acceptable. These factors are
called the acoustic and language models, respectively.

We use the Japanese dictation
toolkit\footnote{http://winnie.kuis.kyoto-u.ac.jp/dictation/}, which
includes the Julius decoder and acoustic/language models. Julius
performs a two-pass (forward-backward) search using word-based forward
bigrams and backward trigrams.  The acoustic model was produced from
the ASJ speech database, which contains approximately 20,000 sentences
uttered by 132 speakers including both gender groups. A 16-mixture
Gaussian distribution triphone Hidden Markov Model, in which states
are clustered into 2,000 groups by a state-tying method, is used.  We
adapt the provided acoustic model by means of an MLLR-based
unsupervised speaker adaptation method, for which we use the HTK
toolkit\footnote{http://htk.eng.cam.ac.uk/}.

Existing methods to adapt language models can be classified into two
categories. In the first category---the {\em integration\/}
approach---general and topic-specific corpora are integrated to
produce a language
model~\cite{auzanne:riao-2000,seymore:eurospeech-97}. Because the
sizes of those corpora differ, N-gram statistics are calculated using
the weighted average of the statistics extracted independently from
those corpora. However, it is difficult to determine the optimal
weight depending on the topic.  In the second category---the {\em
  selection\/} approach---a topic-specific subset is selected from a
general corpus and is used to produce a language model. This approach
is effective if a general corpus contains documents associated with a
target topic, but the N-gram statistics in those documents are
overshadowed by the other documents in the resultant language model.

We followed the selection approach, because the 10M Web page
corpus~\cite{eguchi:sigir-2002} containing mainly Japanese pages
associated with various topics was publicly available.  In practice,
we performed a preprocessing to discard extraneous pages, such as
pages including only word lists and script codes. The resultant corpus
consists of 7M pages.

The quality of the selection approach depends on the method of
selecting topic-specific subsets. An existing
method~\cite{chen:adaptation_ws-2001} uses hypotheses in the initial
speech recognition phase as a query to retrieve topic-specific
documents from a general corpus. However, errors in the initial
hypotheses have the potential to decrease the retrieval
accuracy. Thus, we use the textbook related to a target lecture as a
query to improve the retrieval accuracy and consequently the quality
of the language model adaptation.

\subsection{Document Retrieval}
\label{subsec:retrieval}

We use an existing probabilistic text retrieval
method~\cite{robertson:sigir-94} to compute the relevance score
between the query, which is the textbook for a target lecture, and
each document in the Web corpus. The relevance score for document $d$
is computed by Equation~\eq{eq:okapi}.
\begin{equation}
  \footnotesize
  \label{eq:okapi}
  \sum_{t} f_{t,q}\cdot\frac{\textstyle (K+1)\cdot
  f_{t,d}}{K\cdot\{(1-b)+\textstyle\frac{\textstyle
  dl_{d}}{\textstyle b\cdot avgdl}\} +
  f_{t,d}}\cdot\log\frac{\textstyle N - n_{t} + 0.5}{\textstyle
  n_{t} + 0.5}
\end{equation}
Here, $f_{t,q}$ and $f_{t,d}$ denote the frequency with which term $t$
appears in query $q$ and document $d$, respectively. $N$ and $n_{t}$
denote the total number of documents in the Web corpus and the number
of documents containing term $t$, respectively. $dl_{p}$ denotes the
length of document $d$, and $avgdl$ denotes the average length of
documents in the Web corpus. We empirically set $K=2.0$ and $b=0.8$,
respectively. We use content words, such as nouns, extracted from
transcribed documents as index terms, and perform word-based
indexing. We use the ChaSen morphological
analyzer\footnote{http://chasen.aist-nara.ac.jp/} to extract content
words. The same method is used to extract terms from queries.

\section{Experimentation}
\label{sec:experimentation}

\subsection{Methodology}
\label{subsec:ex_method}

To evaluate the effects of adaptation methods on speech recognition,
we reused the test collection for our previous
work~\cite{fujii:eurospeech-2003}. Five lecture programs on TV, for
which printed textbooks were also published, were videotaped in DV and
were used as target lectures. Each lecture was transcribed manually
and the sentence boundaries with temporal information were also
identified manually.

Table~\ref{tab:test_collection} shows details of the five lectures.
Each lecture was 45 minutes long. We shall use the term ``word token''
to refer to occurrences of words, and the term ``word type'' to refer
to vocabulary items. The column ``\#Fillers'' denoting the number of
interjections in speech partially shows the fluency of each lecturer.

To evaluate the accuracy of speech recognition, we used the word error
rate (WER), which is the ratio of the number of word errors (deletion,
insertion, and substitution) to the total number of words. We also
used test-set out-of-vocabulary rate (OOV) and trigram test-set
perplexity (PP) to evaluate the extent to which our language model
adapted to the target topics.

We used human transcriptions as test set data.  For example, OOV is
the ratio of the number of word tokens not contained in the language
model for speech recognition to the total number of word tokens in the
transcription. Note that smaller values of OOV, PP, and WER are
obtained with better methods.

To adapt language models, we used the textbook for a target lecture
and searched the 7M Web page corpus (see Section~\ref{subsec:asr}) for
the $N$ relevant pages, which were used as a source corpus. We set
$N=5000$, with which the best performance was obtained in a
preliminary experiment.  In the case where the language model
adaptation was not performed, all 7M pages were used as a source
corpus.  We used three different vocabulary sizes, i.e., 20K, 60K, and
100K. In either case, high frequency words in a source corpus were
used to produce a word-based trigram language model. We used the
ChaSen morphological analyzer to extract words from the source
corpora, because Japanese sentences lack lexical segmentation.

For lecture~\#2 we did not perform acoustic model adaptation, because
the speech data contained constant background noise and the sound
quality was not good enough to adapt the acoustic model.

\subsection{Results}
\label{subsec:results}

\begin{table*}[t]
  \begin{center}
    \caption{Details of the five lectures used for experiments.}
    \medskip
    \leavevmode
    \small
    \begin{tabular}{|l|c|c|c|c|c|} \hline
      Lecture ID &
      {\hfill\centering \#1\hfill} &
      {\hfill\centering \#2\hfill} &
      {\hfill\centering \#3\hfill} &
      {\hfill\centering \#4\hfill} &
      {\hfill\centering \#5\hfill}
      \\ \hline\hline
      Topic
      & ~~~Criminal law~~~
      & ~~~Greek history~~~
      & Domestic relations
      & ~~~Food and body~~~
      & ~~~Solar system~~~ \\
      \hline
      \#Word tokens & 6800 & 8040 & 7453 & 8101 & 8235 \\
      \hline
      \#Word types & 1035 & 1223 & 1026 & 905 & 929 \\
      \hline
      \#Sentences & 181 & 191 & 231 & 310 & 340 \\
      \hline
      \#Fillers & 3 & 953 & 818 & 708 & 1134 \\
      \hline
    \end{tabular}
    \label{tab:test_collection}
    
    \bigskip

     \caption{Experimental results for speech recognition (OOV:
       test-set out-of-vocabulary rate (\%), PP: trigram test-set
       perplexity, WER: word error rate (\%)).}
    \medskip
    \leavevmode
    \small
    \tabcolsep=5pt
    \begin{tabular}{|c|l|c|c|c|c|c|c|c|c|c|c|c|c|} \hline
      & & \multicolumn{4}{c|}{20K vocabulary}
      & \multicolumn{4}{c|}{60K vocabulary}
      & \multicolumn{4}{c|}{100K vocabulary} \\
      \cline{3-6} \cline{7-10} \cline{11-14}
      Lecture & & Base & +AM & +LM & +AL
      & Base & +AM & +LM & +AL
      & Base & +AM & +LM & +AL \\ \hline \hline
      & OOV & 
      \multicolumn{2}{c|}{4.83} &
      \multicolumn{2}{c|}{1.19} &
      \multicolumn{2}{c|}{2.06} &
      \multicolumn{2}{c|}{0.54} &
      \multicolumn{2}{c|}{1.56} &
      \multicolumn{2}{c|}{0.54} \\
      \cline{2-14}
      \#1 & PP &
      \multicolumn{2}{c|}{53.62} &
      \multicolumn{2}{c|}{42.16} &
      \multicolumn{2}{c|}{58.73} &
      \multicolumn{2}{c|}{45.99} &
      \multicolumn{2}{c|}{59.00} &
      \multicolumn{2}{c|}{45.99} \\
      \cline{2-14}
      & WER & 31.53 & 21.76 & 21.15 & 12.97 &
      28.16 & 18.38 & 21.24 & 12.84 &
      28.00 & 17.47 & 21.24 & 12.84 \\
      \hline
      & OOV &
      \multicolumn{2}{c|}{9.97} &
      \multicolumn{2}{c|}{7.03} &
      \multicolumn{2}{c|}{3.38} &
      \multicolumn{2}{c|}{1.15} &
      \multicolumn{2}{c|}{2.81} &
      \multicolumn{2}{c|}{0.91} \\
      \cline{2-14}
      \#2 & PP &
      \multicolumn{2}{c|}{116.97} &
      \multicolumn{2}{c|}{105.70} &
      \multicolumn{2}{c|}{203.11} &
      \multicolumn{2}{c|}{236.91} &
      \multicolumn{2}{c|}{209.80} &
      \multicolumn{2}{c|}{240.11} \\
      \cline{2-14}
      & WER & 53.22 & 53.22 & 43.88 & 43.88 & 51.55 & 51.55 & 43.91
      & 43.91 & 51.00 & 51.00 & 43.88 & 43.88 \\
      \hline
      & OOV &
      \multicolumn{2}{c|}{5.44} &
      \multicolumn{2}{c|}{3.86} &
      \multicolumn{2}{c|}{2.55} &
      \multicolumn{2}{c|}{1.56} &
      \multicolumn{2}{c|}{2.19} &
      \multicolumn{2}{c|}{1.56} \\
      \cline{2-14}
      \#3 & PP &
      \multicolumn{2}{c|}{177.73} &
      \multicolumn{2}{c|}{156.42} &
      \multicolumn{2}{c|}{209.87} &
      \multicolumn{2}{c|}{202.37} &
      \multicolumn{2}{c|}{214.21} &
      \multicolumn{2}{c|}{202.37} \\
      \cline{2-14}
      & WER & 69.66 & 59.93 & 61.26 & 53.29 & 69.36 & 59.34 & 61.86
      & 53.53 & 69.27 & 58.80 & 61.86 & 53.53 \\
      \hline
      & OOV &
      \multicolumn{2}{c|}{6.43} &
      \multicolumn{2}{c|}{3.12} &
      \multicolumn{2}{c|}{2.49} &
      \multicolumn{2}{c|}{2.67} &
      \multicolumn{2}{c|}{2.04} &
      \multicolumn{2}{c|}{2.67} \\
      \cline{2-14}
      \#4 & PP &
      \multicolumn{2}{c|}{126.69} &
      \multicolumn{2}{c|}{117.76} &
      \multicolumn{2}{c|}{165.08} &
      \multicolumn{2}{c|}{123.28} &
      \multicolumn{2}{c|}{169.89} &
      \multicolumn{2}{c|}{123.28} \\
      \cline{2-14}
      & WER & 62.30 & 52.50 & 48.86 & 39.32 & 59.56 & 48.48 & 48.90
      & 39.86 & 58.07 & 47.72 & 48.90 & 39.86 \\
      \hline
      & OOV &
      \multicolumn{2}{c|}{7.33} &
      \multicolumn{2}{c|}{4.16} &
      \multicolumn{2}{c|}{2.14} &
      \multicolumn{2}{c|}{0.47} &
      \multicolumn{2}{c|}{1.77} &
      \multicolumn{2}{c|}{0.47} \\
      \cline{2-14}
      \#5 & PP &
      \multicolumn{2}{c|}{186.89} &
      \multicolumn{2}{c|}{125.36} &
      \multicolumn{2}{c|}{266.45} &
      \multicolumn{2}{c|}{191.94} &
      \multicolumn{2}{c|}{272.74} &
      \multicolumn{2}{c|}{191.94} \\
      \cline{2-14}
      & WER & 78.83 & 64.98 & 58.26 & 48.97 & 77.62 & 63.90 & 58.77
      & 48.77 & 76.87 & 63.58 & 58.77 & 48.77 \\
      \hline
      & OOV &
      \multicolumn{2}{c|}{7.42} &
      \multicolumn{2}{c|}{4.16} &
      \multicolumn{2}{c|}{2.61} &
      \multicolumn{2}{c|}{1.34} &
      \multicolumn{2}{c|}{2.14} &
      \multicolumn{2}{c|}{1.29} \\
      \cline{2-14}
      Avg. & PP &
      \multicolumn{2}{c|}{132.38} &
      \multicolumn{2}{c|}{109.48} &
      \multicolumn{2}{c|}{180.65} &
      \multicolumn{2}{c|}{160.10} &
      \multicolumn{2}{c|}{185.13} &
      \multicolumn{2}{c|}{160.74} \\
      \cline{2-14}
      & WER & 59.93 & 50.83 & 47.34 & 39.46 & 58.10 & 48.58 & 47.59
      & 39.57 & 57.47 & 47.96 & 47.58 & 39.57 \\
      \hline
    \end{tabular}
    \label{tab:results}
  \end{center}
\end{table*}

Table~\ref{tab:results} shows the values of OOV, PP, and WER for
different methods. The columns ``Base'', ``+AM'', ``+LM'', and ``+AL''
denote the results obtained by no adaptation, acoustic model
adaptation, language model adaptation, and acoustic/language model
adaptation, respectively.

Note that the values of OOV and PP do not change whether or not the
acoustic model adaptation was performed. It should also be noted that
because for lecture~\#2 the acoustic model adaptation was not
performed, the values of WER for +AM and +AL are the same as those for
Base and +LM, respectively.

Suggestions which can be derived from Table~\ref{tab:results} are as
follows.
First, the values of OOV and PP decreased by language model
adaptation, excepting lectures~\#2 and \#3 with 20K and 60K vocabulary
sizes.

Second, the values of WER generally decreased by adapting acoustic and
language models, independently. The improvement obtained by each model
was almost the same. However, when used together the improvement was
even greater, resulting in the average value of WER was less than
40\%.  This result is encouraging because in the TREC spoken document
retrieval track~\cite{garofolo:trec-97}, the decrease of the retrieval
accuracy was small if the value of WER was 30-40\%.

Third, the values of WER decreased by adapting acoustic and language
models even for the 60K and 100K vocabulary sizes, although in our
previous study the vocabulary size was limited to 20K.

However, the values of WER themselves did not change significantly
depending on the vocabulary size.  One reason is that the average
number of word types  (i.e., the actual vocabulary size) in the 5,000
pages used to adapt a language model was approximately 54K, which was
less than the permissible number. However, in the cases where we used
all 7M pages, the vocabulary size was always the same as the
permissible number.

Finally, by comparing the values of WER for 20K vocabulary size
obtained with the adaptation (+AM, +LM, and +AL) and those for 60K and
100K vocabulary sizes obtained without the adaptation (Base), one can
see that adapting the language model and/or the acoustic model was
more effective than increasing the vocabulary size.

\section{Conclusion}
\label{sec:conclusion}

In this paper, to improve the quality of cross-media information
retrieval system for lecture videos, we evaluated the effects of
adapting acoustic and language models on speech recognition.  We
performed an MLLR-based method to adapt an acoustic model. To obtain a
corpus for language model adaptation, we used the textbook for a
target lecture to search a Web collection for the pages associated
with the lecture topic.  The experimental results for five lectures
showed that  the methods to adapt acoustic and language models
improved the speech recognition accuracy independently and when used
together the improvement was even greater.

\small
\bibliographystyle{IEEEtran}

\begin{thebibliography}{1}
\providecommand{\url}[1]{#1}
\def\UrlFont{\rmfamily}
\providecommand{\newblock}{\relax}
\providecommand{\bibinfo}[2]{#2}
\providecommand\BIBentrySTDinterwordspacing{\spaceskip=0pt\relax}
\providecommand\BIBentryALTinterwordstretchfactor{4}
\providecommand\BIBentryALTinterwordspacing{\spaceskip=\fontdimen2\font plus
\BIBentryALTinterwordstretchfactor\fontdimen3\font minus
  \fontdimen4\font\relax}
\providecommand\BIBforeignlanguage[2]{{%
\expandafter\ifx\csname l@#1\endcsname\relax
\typeout{** WARNING: IEEEtran.bst: No hyphenation pattern has been}%
\typeout{** loaded for the language `#1'. Using the pattern for}%
\typeout{** the default language instead.}%
\else
\language=\csname l@#1\endcsname
\fi
#2}}

\bibitem{fujii:eurospeech-2003}
A.~Fujii, K.~Itou, T.~Akiba, and T.~Ishikawa, ``A cross-media retrieval system
  for lecture videos,'' in \emph{Proceedings of the 8th European Conference on
  Speech Communication and Technology}, 2003, pp. 1149--1152.

\bibitem{auzanne:riao-2000}
C.~Auzanne, J.~S. Garofolo, J.~G. Fiscus, and W.~M. Fisher, ``Automatic
  language model adaptation for spoken document retrieval,'' in
  \emph{Proceedings of RIAO 2000 Conference on Content-Based Multimedia
  Information Access}, 2000.

\bibitem{seymore:eurospeech-97}
K.~Seymore and R.~Rosenfeld, ``Using story topics for language model
  adaptation,'' in \emph{Proceedings of Eurospeech97}, 1997, pp. 1987--1990.

\bibitem{eguchi:sigir-2002}
K.~Eguchi, K.~Oyama, K.~Kuriyama, and N.~Kando, ``The {Web} retrieval task and
  its evaluation in the third {NTCIR} workshop,'' in \emph{Proceedings of the
  25th Annual International ACM SIGIR Conference on Research and Development in
  Information Retrieval}, 2002, pp. 375--376.

\bibitem{chen:adaptation_ws-2001}
L.~Chen, J.-L. Gauvain, L.~Lamel, G.~Adda, and M.~Adda, ``Language model
  adaptation for broadcast news transcription,'' in \emph{Proceedings of ISCA
  Workshop on Adaptation Methods For Speech Recognition}, 2001.

\bibitem{robertson:sigir-94}
S.~Robertson and S.~Walker, ``Some simple effective approximations to the
  2-poisson model for probabilistic weighted retrieval,'' in \emph{Proceedings
  of the 17th Annual International ACM SIGIR Conference on Research and
  Development in Information Retrieval}, 1994, pp. 232--241.

\bibitem{garofolo:trec-97}
J.~S. Garofolo, E.~M. Voorhees, V.~M. Stanford, and K.~S. Jones, ``{TREC-6}
  1997 spoken document retrieval track overview and results,'' in
  \emph{Proceedings of the 6th Text REtrieval Conference}, 1997, pp. 83--91.

\end{thebibliography}

\end{document}